\documentclass[authoryear,preprint,12pt]{elsarticle}

\usepackage{xcolor}
\usepackage{color,soul}

\usepackage{amssymb}
\usepackage{subcaption}

\usepackage{url}

\usepackage{booktabs} % For professional looking tables
\usepackage{multirow}

\journal{Natrual Language Processing}

\begin{document}

\begin{frontmatter}


\author[1]{Rrubaa Panchendrarajan\corref{cor1}}
\ead{r.panchendrarajan@qmul.ac.uk}

\author[1]{Arkaitz Zubiaga}
\ead{a.zubiaga@qmul.ac.uk}
\ead[url]{www.zubiaga.org}

\cortext[cor1]{Corresponding author}

\title{Claim Detection for Automated Fact-checking: A Survey on Monolingual, Multilingual and Cross-Lingual Research}

%\author[]{}
\affiliation[1]{organization={School of Electronic Engineering and Computer Science, Queen Mary University of London},
             addressline={327 Mile End Rd, Bethnal Green},
             city={London},
             postcode={E1 4NS},
             %state={},
             country={United Kingdom}}

\begin{abstract}
Automated fact-checking has drawn considerable attention over the past few decades due to the increase in the diffusion of misinformation on online platforms. This is often carried out as a sequence of tasks comprising (i) the detection of sentences circulating in online platforms which constitute claims needing verification, followed by (ii) the verification process of those claims. This survey focuses on the former, by discussing existing efforts towards detecting claims needing fact-checking, with a particular focus on multilingual data and methods. This is a challenging and fertile direction where existing methods are yet far from matching human performance due to the profoundly challenging nature of the issue. Especially, the dissemination of information across multiple social platforms, articulated in multiple languages and modalities demands more generalized solutions for combating misinformation. Focusing on multilingual misinformation, we present a comprehensive survey of existing multilingual claim detection research. We present state-of-the-art multilingual claim detection research categorized into three key factors of the problem, verifiability, priority, and similarity. Further, we present a detailed overview of the existing multilingual datasets along with the challenges and suggest possible future advancements.  
\end{abstract}


%Research highlights
%\item Comprehensive review of multilingual claim detection
%\item Discuss claim detection task categorized into verifiability, priority, and similarity identification.
%\item Overview of state-of-the-art approaches for multilingual claim detection
%\item Existing multilingual datasets for a wide range of claim detection tasks 
%\item Open challenges and future directions of multilingual claim detection

\begin{keyword}
Multilingual Claim Detection \sep Claim Prioritization \sep Claim Matching \sep Claim Clustering \sep Fact-checking



\end{keyword}

\end{frontmatter}


\section{Introduction}
Misinformation poses a significant threat to society, a threat that has escalated with the advent and widespread use of social media platforms. This is demanding an additional layer of verifying online information to ensure the integrity and validity of the information that people read online. However, verifying the content circulating on online platforms is a time-consuming task that if done manually can only encompass a small portion of the available information, which demands the development of methods to enable automated fact-checking, a process that starts off by identifying information needing verification and ends up by verifying whether a claim is supported or refuted by a reputable piece of evidence, or occasions with a verdict that there is no sufficient evidence to determine its accuracy. This is often carried out as a series of steps involving (i) the identification of claims to be checked, (ii) prioritization of important claims to deal with, (iii) gathering evidence associated with those claims, and (iv) concluding with the final verdict by checking the claim against the associated evidence. While there are several dedicated organizations such as PoliFact\footnote{\url{https://www.politifact.com/}}, Full Fact\footnote{\url{https://fullfact.org/}} and Newtral\footnote{\url{https://www.newtral.es/}} established in recent years, research in this direction is experiencing a substantial increase in both fact-checking organizations and academic research due to the growing pressure of dealing with online misinformation. 

There are some recent surveys presenting overviews of existing research on the fact-checking pipeline and its underlying components \citep{zeng2021automated,guo2022survey,das2023state}. However, their holistic focus on the entire fact-checking process impedes them from providing a detailed study of each component of the pipeline. In addition to these surveys, some studies have focused on reviewing each a particular aspect of the fact-checking problem. For example, \cite{hardalov2022survey} highlights the role of \textit{stance detection} in automated misinformation detection, and \cite{kotonya2020explainable,kotonya2020explainable_automated} focus on the explainability aspect of automated fact-checking. Different from these survey papers, we present a comprehensive study on the claim detection component of the automated fact-checking pipeline, with a specific focus on multilingual research. 

\begin{figure}[t!]
\centering    
\includegraphics[width=1\textwidth]{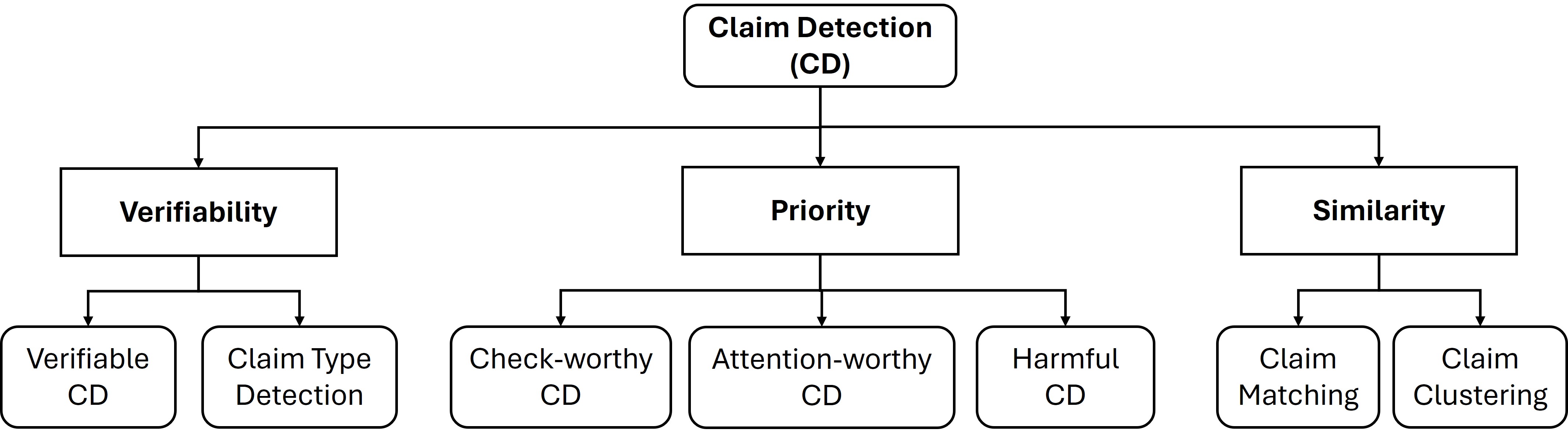}
\caption{Claim detection tasks.}
\label{fig:Claim_Detection_Tasks}
\end{figure}

This survey presents a comprehensive review of the state-of-the-art techniques used for a wide range of claim detection tasks. Figure \ref{fig:Claim_Detection_Tasks} depicts the claim detection tasks discussed in this paper. Given that the claim detection task can have different objectives and hence be formulated in different ways, we discuss the different claim detection subtasks by grouping them into the following three categories:
\begin{itemize}
    \item \textbf{Verifiability}: Identifying claims that are verifiable. We further discuss the definition of verifiable clams, and the tasks associated with it in Section \ref{sec:background}.
    \item \textbf{Priority}: Not all the verifiable claims are worthy of fact-checking, and prioritization of claims plays a vital role in effective fact-checking. We further discuss the factors determining the priority of claims, and the tasks associated with it in Section \ref{sec:background}.
    \item \textbf{Similarity}: A massive amount of unverified online content often comprises repeated information. Hence, identifying similar claims is important for avoiding the repetition of fact-checking similar claims. We introduce the similarity identification tasks in Section \ref{sec:background}.
\end{itemize}

The rest of the survey is structured as follows. Section \ref{sec:background} introduces the fact-checking pipeline, different definitions used in the literature to define a claim, and the multilingual view of the claim detection problem. Sections \ref{sec:verifiable claim detection} and \ref{sec:claim_prioritization} present existing research on identifying \textit{verifiability} and \textit{priority} of claims. Similarity identification of the claims is discussed in Sections \ref{sec:claim matching} and \ref{sec:claim clustering}. We outline the challenges associated with claim detection in Section \ref{sec:challenges}, followed by the conclusions in Section \ref{sec:conclusion}. 

\section{Background: Automated Fact-checking Pipeline}\label{sec:background}

% \subsection{Automated Fact-checking Pipeline}\label{sec:fact-checking_background}

\begin{figure}[htbp!]
\centering    
\includegraphics[width=1\textwidth]{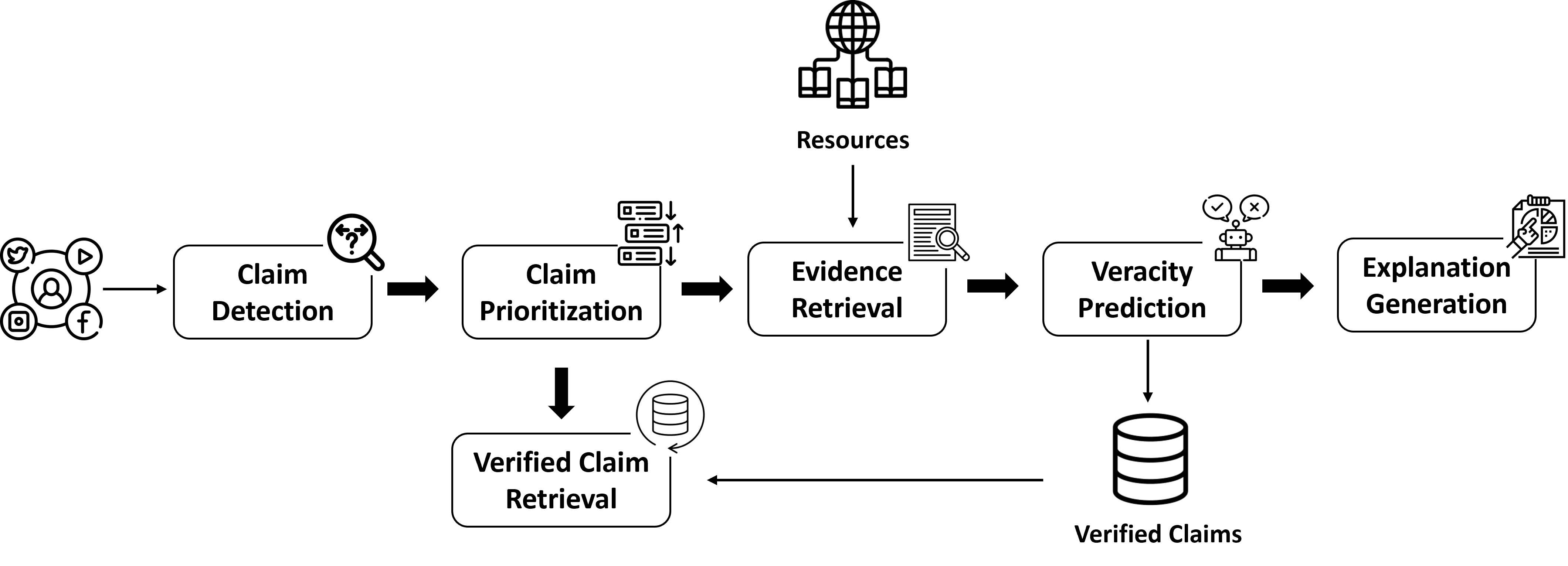}
\caption{Fact-checking Pipeline}
\label{fig:Fact-checking Pipeline}
\end{figure}

An automated fact-checking pipeline is typically composed of five major components \citep{das2023state} as depicted in Figure \ref{fig:Fact-checking Pipeline}. The process begins with detecting, out of a collection of input sentences, verifiable factual statements referred to as \textit{claims}. This component is commonly applied to social platforms and online resources such as news articles to identify statements that require verification, hence getting rid of the remainder of the sentences which do not need to go through the rest of the fact-checking pipeline for not needing verification. Identifying the verifiability of claims is often carried out as a binary decision indicating whether a statement is verifiable or not. However, determining the claim type indicating the type of factual information presented in the claim can also be performed as a fine-grained analysis, as for example \cite{konstantinovskiy2021toward} introduced a taxonomy of types of claims.  

Once the claims are extracted, they go through a prioritization process to estimate the worthiness of the claim to be verified. The criteria used to estimate the worthiness may vary according to the topic or domain of the claim and the user groups (i.e. audience) who are interested in the veracity of the claim. Some of the popular criteria used in the literature are the virality of the claim, the interest of the public in the veracity of the claim, the impact that the claim could make on society, and its timeliness \citep{das2023state,micallef2022true}. While the determination of priority is often modeled as the estimation of the \textit{check-worthiness} of a claim, similar prioritization tasks including estimation of \textit{attention-worthiness} and \textit{harmfulness} are also carried out in the literature \citep{nakov2022overview} for claim prioritization.

Following the prioritization, relevant evidence is retrieved which could ideally help support or refute the prioritized claims, and finally, the verdict of the claim indicating whether the fact discussed in the statement is supported or refuted\footnote{Occasionally some researchers may choose to output `true' or `false' predictions, although `supported' and `refuted' are the more widely used alternatives indicating how the retrieved evidence aligns with the claim, rather than concluding the veracity of the claim.} is predicted with respect to the evidence retrieved \citep{guo2022survey}, while also on alternative occasions predicting that there is `not enough information' to determine the support or refutal. Often the evidence retrieval and veracity prediction tasks are tackled jointly in the literature and referred to as the \textit{fact verification} process \citep{guo2022survey}. The latest addition to the automated fact-checking pipeline is the explanation generation \citep{kotonya2020explainable}, where researchers aim to automatically generate a reason for the verdict predicted to boost the interpretability and explainability of the system. 

Apart from these five major components, as a parallel component, researchers have focused on retrieving, from the database of previously fact-checked claims, existing claims that resemble or relate to newly retrieved claims. This task is referred to as \textit{verified claim retrieval} or \textit{claim matching} in the literature. Matching new claims with previously fact-checked claims can then help avoid a repeated attempt of claim verification as the verdict was already done in the past. While this can avoid the substantial time involved in processing an unverified claim through the remaining components of the pipeline, the impact and spread of the claim can also be minimized with timely verdicts.   

\subsection{Definition of Claim}

\begin{figure}[t!]
\centering    
\includegraphics[width=1\textwidth]{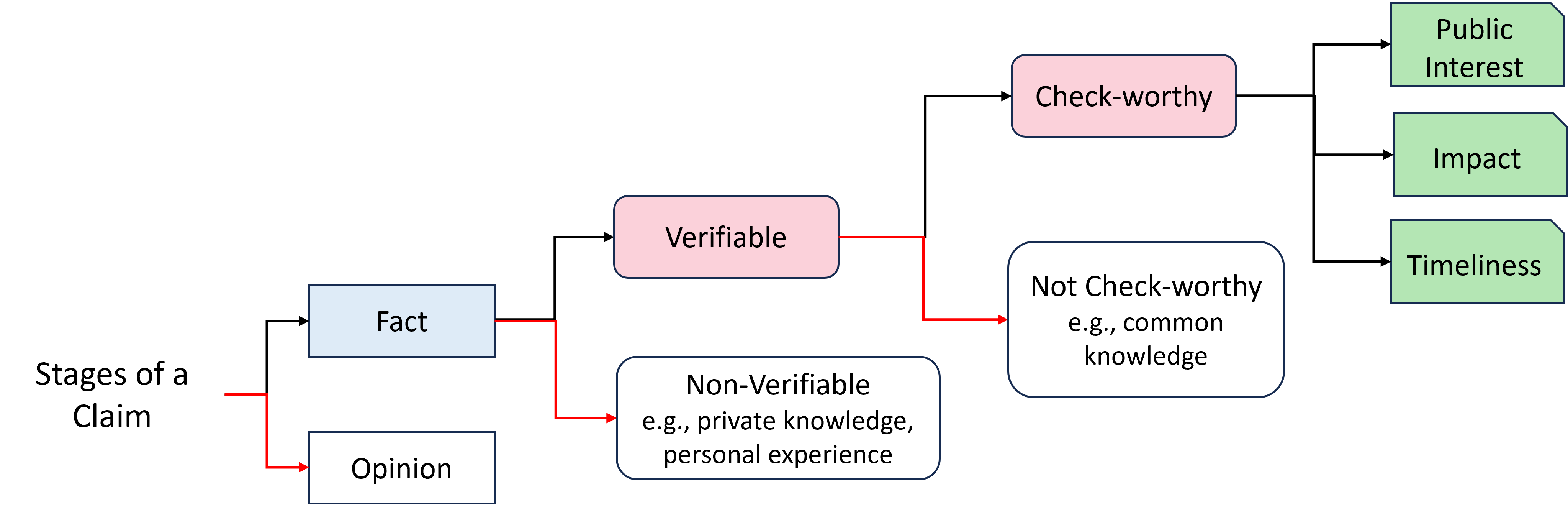}
\caption{Stages of a claim}
\subcaption*{Black arrows in the figure indicate claims and red arrows indicate non-claims}
\label{fig:Claim_defnition}
\end{figure}

There are different ways of defining a claim depending on the objective of different components of the automated fact-checking pipeline. This includes the following three key stages of a claim as depicted in Figure \ref{fig:Claim_defnition}:
\begin{itemize}
    \item \textit{Claim} - A claim is defined as a statement containing a purported fact about the real world \citep{das2023state}. For example, the statement \textit{``team X is the best team in FIFA 2023"} is an opinion, whereas, the statement \textit{``United States is one of the host countries of the next FIFA World Cup"} is a claim that can be checked against an objective piece of evidence. \cite{zeng2021automated} give a slightly different definition for a claim as a ``factual statement under investigation".   
    
    \item \textit{Verifiable Claim} - A verifiable claim is defined as a ``factual statement that can be checked" by \cite{micallef2022true}, and a similar definition of ``assertion about the world that is checkable" is given by \cite{konstantinovskiy2021toward}. Claims about personal experience are neither an assertion about the world nor can be verified. Therefore they are not considered as \textit{verifiable claims}. While the definition of a \textit{verifiable claim} enforces the possibility of determining the veracity of a claim, this aspect predominantly relies on the availability of evidence. However, the availability of evidence can not be determined until the execution of the evidence retrieval task. Therefore, verifiable claim detection tasks often ignore the availability of evidence and aim at identifying claims about the real world.
    
    \item \textit{Check-worthy Claim} - Verifying all the assertions about the real world is impractical, and hence it demands a prioritization process. This objective is generally handled as an estimation of check-worthiness of a verifiable claim in the literature. Due to the subjective nature of this task, it is hard to define the check-worthiness of a verifiable claim and it may vary according to the topic discussed in the claim and the user group who are interested in the claim. Further, the worthiness may vary over time \citep{guo2022survey}, as the interest of a claim may fade or increase as stories develop, which makes it more challenging. Some of the common factors used to determine the check-worthiness of verifiable claims include the popularity of the claim, amount of public interest in the verdict of the claim, impact of the verdict, and timeliness of the verdict \citep{das2023state,micallef2022true}. In addition to the check-worthiness, the following criteria are used in the literature for prioritizing claims:
    \begin{itemize}
        \item \textit{Attention-worthy Claims}: A verifiable claim that should get the attention of the policymakers and government entities \citep{nakov2022overview,NLP4IF-2021-COVID19-task}.
        \item \textit{Harmful Claims}: A verifiable claim that is harmful to society \citep{nakov2022overview,NLP4IF-2021-COVID19-task}.
        \item \textit{Interesting to the general public}: A verifiable claim that may have an impact on society or attract interest from the general public \citep{NLP4IF-2021-COVID19-task}.
    \end{itemize}
\end{itemize}

Apart from these three key stages of a claim, \textit{rumor} detection is also typically handled as a similar problem in the literature, although not strictly formulated as a fact-checking task, where a statement can be categorized as a rumor or non-rumor. A rumor is defined in the literature \cite{zubiaga2018detection} as ``an unverified claim circulating on social platforms" \citep{guo2022survey} and as ``unverified and instrumentally relevant information statements in circulation” \citep{difonzo2007rumor}.   

\subsection{Cross-lingual Claim Detection}

\begin{figure}[t!]
\centering    
\includegraphics[width=1\textwidth]{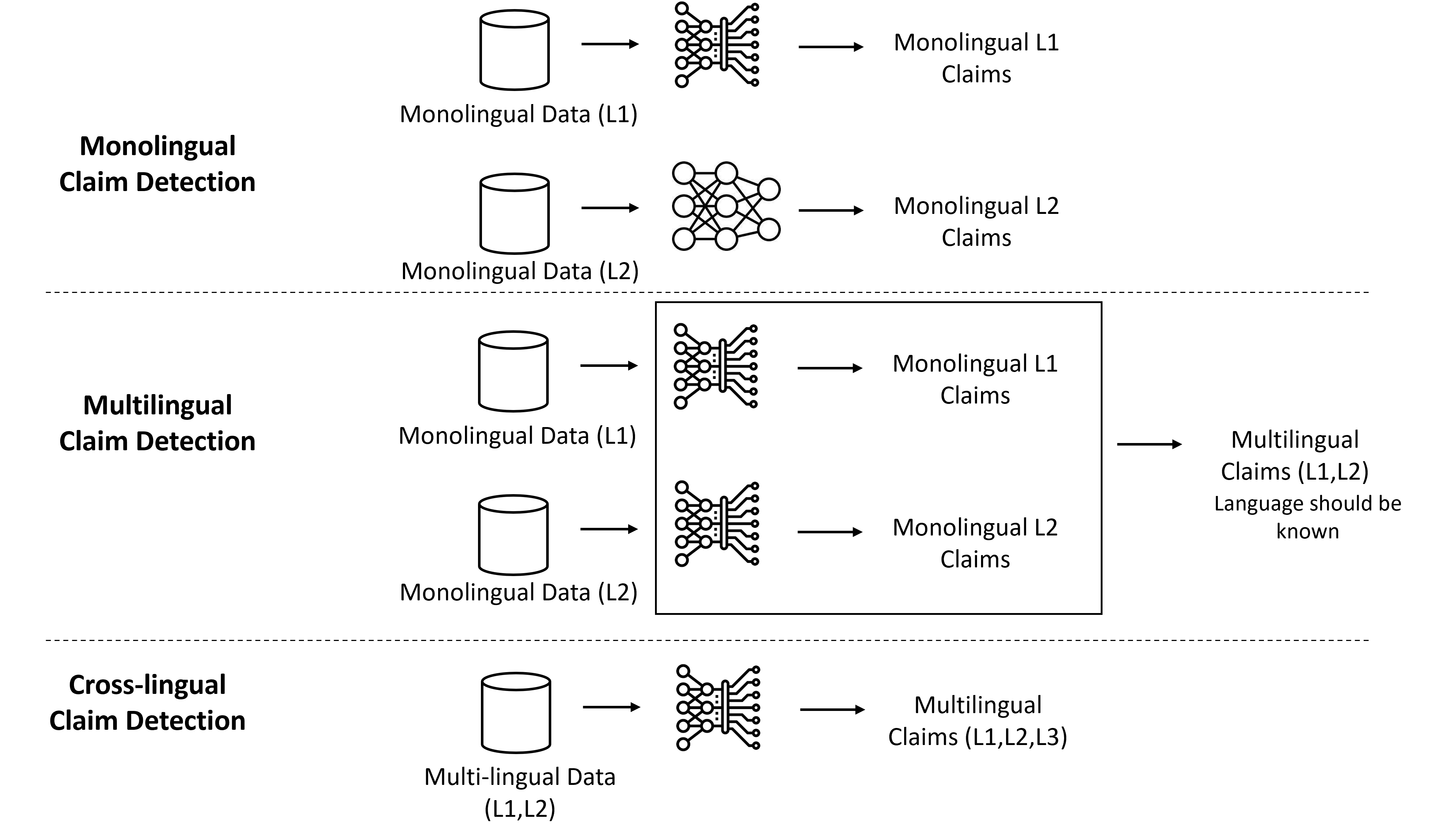}
\caption{Monolingual vs Multilingual vs Cross-lingual Claim Detection.}
\label{fig:Multi-lingual_Explanation}
\end{figure}

We define the following terms used in the literature to refer to the multilingual aspect of the claim detection problem. Figure \ref{fig:Multi-lingual_Explanation} summarizes these definitions.
\begin{itemize}
    \item \textit{Monolingual} Claim Detection - Achieved via developing a language-specific model for identifying the claims of a specific language \textit{L1}, and training the model on the data from the same language. Repeating the process for language \textit{L2} remains as \textit{monolingual} claim detection as the introduction of new language \textit{L2} requires developing a language-specific model.   
    
    \item \textit{Multilingual} Claim Detection - Achieved via developing a language-independent claim detection model, and training it on the data from language \textit{L1}. Applying the model for other languages requires retraining the model for other languages on language-specific training data. Therefore, this approach demands    
    training data from each language the fact-checking pipeline deals with and the language of the input statement to be known to determine the corresponding \textit{monolingual} model. Further, the \textit{monolingual} models do not generalize their knowledge beyond the language it has seen in the training data.  

    \item \textit{Cross-lingual} Claim Detection - Achieved via developing a single model for identifying the claims of any language or a broader set of languages than the one(s) seen during training. The \textit{cross-lingual} model is trained using training data composed of one or more languages. Further, it generalizes its knowledge beyond the languages it has seen in the training data, to identify claims written in new languages. 
\end{itemize}

\subsection{Transformer Models}

The transformer architecture \citep{vaswani2017attention} proposed in 2017 was a breakthrough in the evolution of neural architectures. The model was composed of layers of encoders and decoders with multi-headed attention, enabling them to be very effective and powerful in sequence transduction tasks. Later, the architecture was explored by researchers to introduce several influential models, resulting in a family of transformer models. Some of the notable models include BERT (Bidirectional Encoder Representations from Transformers) \citep{bert}, GPT (Generative Pre-trained Transformer) \citep{GPT}, BART (Bidirectional Auto-Regressive Transformers) \citep{bart}, and their variations. 

Due to the powerful nature of these models in understanding the language and training data, fine-tuning them with limited training data showed a significant performance increase in several Natural Language Processing (NLP) tasks \citep{pfeiffer2020adapterhub}. This generalization capability of the transformer models was further extended to multi-lingual settings, by training them on massive amounts of multi-lingual data. mBERT, mGPT, mBART, and XLM-R \citep{xlm-r} are some of the popular multi-lingual pre-trained transformer models, which offer state-of-the-art performance in cross-lingual settings in various NLP tasks.

\section{Verifiable Claim Detection}\label{sec:verifiable claim detection}
This section summarizes the existing verifiable claim detection research in the literature. This task is often treated as a binary classification problem to identify whether a statement is a verifiable claim or not. However, the problem is treated as a multi-class classification too by either adding an uncertainty label \citep{kazemi-etal-2021-claim} or identifying fine-grained verifiable claim types \citep{konstantinovskiy2021toward}. 

\subsection{Datasets}
A key challenge in cross-lingual setting is the availability of multi-lingual training data. Some notable datasets released for multilingual verifiable claim detection were the NLP4IF 2021 shared task data \citep{NLP4IF-2021-COVID19-task} and CheckThat 2022 data \citep{nakov2022overview}. Both datasets contain tweets related to the COVID-19 pandemic. NLP4IF 2021 includes tweets written in English, Arabic, and Bulgarian languages labeled as verifiable claims or not. The annotation also includes six other misinformation labels including the label indicating whether the tweet requires fact-checking or not. In addition to these three languages, the CheckThat 2022 dataset includes Dutch and Turkish tweets labeled for the verifiable claim detection and claim prioritization tasks. This dataset was expanded with more languages and data via several stages \citep{alam2021fighting,shaar2021overview,nakov2021second}, and the final version was released in 2022 \citep{nakov2022overview}. Following these, several other multilingual verifiable claim detection datasets were released focusing on topics including COVID-19 \citep{kazemi-etal-2021-claim} and politics \citep{kazemi-etal-2021-claim,dutta2022multilingual}. Table \ref{table:verifiable_claim_detection_datasets} summarizes the existing multilingual claim detection datasets. While there are more datasets of monolingual nature, only the four datasets listed in the table include multilingual data, with dataset sizes varying from 1.3K to 6K and including a diverse set of languages.

\begin{table}
\caption{Multilingual Verifiable Claim Detection Datasets}
\centering
\label{table:verifiable_claim_detection_datasets}
\footnotesize
\hspace*{-5em}%Added due to space issue
\begin{tabular}{llllllll}
\toprule
Dataset & Objective & Label & Language  & Topic & Source & Size & Evaluation       \\\midrule
\begin{tabular}[c]{@{}l@{}}NLP4IF 2021\\ \citep{NLP4IF-2021-COVID19-task} \end{tabular}                & \begin{tabular}[c]{@{}l@{}}Verifiable\\Claims\end{tabular}         & \begin{tabular}[c]{@{}l@{}}Yes\\ No\end{tabular}            & \begin{tabular}[c]{@{}l@{}}English\\ Arabic   \\ Bulgarian\end{tabular} & Covid-19                            & Twitter         & 1.3K – 4K    & \begin{tabular}[c]{@{}l@{}}Precision\\Recall\\F1 Score\end{tabular} \\ \midrule
\cite{kazemi-etal-2021-claim}            & \begin{tabular}[c]{@{}l@{}}Claim-like\\Statements\end{tabular} & \begin{tabular}[c]{@{}l@{}}Yes\\ No\\ Probably\end{tabular} & \begin{tabular}[c]{@{}l@{}}English\\ Hindi\\ Bengali\\ Malayalam\\ Tamil\end{tabular}        & \begin{tabular}[c]{@{}l@{}}Covid-19\\ Politics\end{tabular} & WhatsApp        & 5K         &  \begin{tabular}[c]{@{}l@{}}Accuracy\\Precision\\Recall\\F1 Score\end{tabular} \\ \midrule
\cite{dutta2022multilingual} & \begin{tabular}[c]{@{}l@{}}Verifiable\\Claims\end{tabular}  & \begin{tabular}[c]{@{}l@{}}Yes\\ No\end{tabular}            & \begin{tabular}[c]{@{}l@{}}English\\ Hindi    \\ Bengali\\ Code-mixed\end{tabular}           & Politics                                                    & Twitter         & 600 – 1.4K &  \begin{tabular}[c]{@{}l@{}}Precision\\Recall\\F1 Score\end{tabular} \\ \midrule
\begin{tabular}[c]{@{}l@{}}CheckThat 2022\\ \citep{nakov2022overview} \end{tabular}                & \begin{tabular}[c]{@{}l@{}}Verifiable\\Claims\end{tabular}         & \begin{tabular}[c]{@{}l@{}}Yes\\ No\end{tabular}            & \begin{tabular}[c]{@{}l@{}}English\\ Arabic   \\ Bulgarian\\ Dutch   \\ Turkish\end{tabular} & Covid-19                            & Twitter         & 4K – 6K    & Accuracy \\
\bottomrule
\end{tabular}
\end{table}

\subsection{Cross-lingual Claim Detection}
One of the pioneering works in cross-lingual claim detection was carried out by the authors of the NLP4IF dataset \citep{alam2020fighting}. They experimented with both cross-lingual and mono-lingual settings using multi-lingual models including mBERT (multilingual BERT) \citep{bert} and XLM-r \citep{xlm-r} and language-specific BERT models. The authors observed similar or improved performance in cross-lingual settings when the models were trained using the combined dataset (dataset including all three languages under study). Further, the authors experimented with the impact of tweet-specific features on the performance. This includes various features of a tweet and its author details such as verified status, number of friends, and followers. Additionally, the botness of the tweet indicating the chances of the author being a bot is also included as a feature for the model. Compared to other features, the authors observed an increase in performance when the botness feature was injected into the classification model. A similar attempt was made by \cite{uyangodage2021can} in experimenting with both cross-lingual and mono-lingual settings using BERT variations. The authors fine-tuned publicly available language-specific BERT models and mBERT \citep{bert} using the language-specific and combined training data respectively. They observed similar or improved performance in cross-lingual settings on the NLP4IF dataset. 

\cite{panda2021detecting} performed a similar analysis of utilizing mBERT \citep{bert} for the classification task in the NLP4IF 2021 dataset, and the authors reported that mBERT can achieve an impressive score in identifying misinformation labels even without fine-tuning on language-specific training data. A recent study by \cite{agrestia2022polimi} showed, that fine-tuning GPT-3 model \citep{GPT} using English data only gives competitive performance to the BERT models trained on language-specific training data for both verifiable claim detection and claim prioritization tasks. 

\subsection{Multilingual Claim Detection}
A straightforward approach to implementing the multi-lingual setting for verifiable claim detection tasks is fine-tuning the multi-lingual pre-trained models on language-specific training data. Following this approach \cite{husunbeyi2022rub} used XLM-R \citep{xlm-r} for training language-specific models in CheckThat 2022 English and Turkish data for both the verifiable claim detection and claim prioritization tasks. The authors compared the performance with monolingual pre-trained models and observed that multilingual models achieve competitive performance when trained on language-specific training data. \cite{savchev2022ai} performed a similar analysis using the XLM-R multilingual model. The author further used back translation, translating a tweet to a target language, and then translating back from the target language to the original language as the data augmentation technique. They observed an increase in overall performance with the incorporation of the data augmentation technique. The latest study by  \cite{eyuboglu2023fight} applied a wide range of pre-trained models and observed that, generally, BERT variants are very powerful in classifying both the verifiability and priority level of claims.    

\subsection{Monolingual Claim Detection}
One of the earliest works in the direction of monolingual verifiable claim detection was carried out by  \cite{prabhakar2020claim}. The authors used the verifiable claims from the dataset FEVER \citep{thorne-etal-2018-fever} and collected the non-claims from Wikipedia articles with the assumption that sentences without any citation are non-claims according to Wikipedia's verifiability policy.\footnote{\url{https://en.wikipedia.org/wiki/Wikipedia:Verifiability}} This resulted in a massive amount of claim and non-claim samples in English. The authors fine-tuned BERT \citep{bert} and DistilBERT \citep{distilbert} models to identify the claims and observed that the fine-tuned BERT model achieved an F1-score of 98\%. Similarly, \cite{alam2021fighting} released the initial version of the CheckThat dataset and the authors experimented with different sets of BERT variations for each language. Among the variations of BERT models, they observed that XLM-r \citep{xlm-r} outperformed other models in several languages.      

Similar to the cross-lingual setting, \cite{suri2022asatya} used the BERT model with a data augmentation technique to detect verifiability and priority of claims in the English language in CheckThat 2022 data. The authors translated the training data from other languages to English to increase the training data size. Further, they injected both tweet-specific and author-specific features as additional input to the model for improving the classification performance. Apart from these studies, language-specific pre-trained BERT models were often used as the effective monolingual solution \citep{henia2021icompass,hussein2021damascusteam}. 

Different from these approaches, \cite{konstantinovskiy2021toward} performed a fine-grained analysis of verifiable claims by classifying a sentence into non-claim or six sub-categories of a claim. The authors annotated around 6300 sentences from subtitles of TV political debates and trained various traditional machine learning models with a wide range of features. Notable textual features include TF-IDF, Part-of-speech (POS) tags, Named entity recognition (NER), and word embeddings. The authors observed that the logistic regression classifier \citep{lavalley2008logistic} obtained the highest F1 score in classifying the sentences as a claim or non-claim, and injecting POS and NER information did not improve the performance of the optimal classifier. The proposed solution was tested in a live feed of transcripts from TV shows.  Similar claim type identification research has been carried out in the literature using rule-based approaches \citep{rony2020claimviz}. 

The most recent attention on monolingual claim detection has been given to developing domain-specific solutions.  \cite{woloszyn2021towards} focused on identifying \textit{green claim}, a claim discussing an issue related to the environment. The authors compared three pre-trained models, RoBERTa \citep{liu2019roberta}, BERTweet \citep{nguyen2020bertweet}, and Flair \citep{akbik2018contextual}, and observed that generally, RoBERTa outperformed the other two models in the green claim detection task. \cite{smeros2021sciclops} extracted scientific claims by introducing three variants of BERT, SciBERT, NewsBERT, and SciNewsBERT fine-tuned using scientific articles and news headlines. Recent research has also shown that the difficulty of identifying check-worthy claims varies across domains. \cite{abumansour2023check} showed that a model trained on check-worthy and non-check-worthy claims pertaining to a set of topics can struggle to perform the claim detection task on a new, unseen topic, a limitation that can be mitigated through the use of data augmentation strategies.

\subsection{Evaluation}\label{sec:verifiable-claim-detection-evaluation}
As previously mentioned, identifying a verifiable claim is consistently approached as a classification task, and typically as a binary classification task. The following evaluation metrics are used to evaluate the classification models in the literature.
\begin{itemize}
    \item Accuracy \citep{nakov2022overview} - Proportion of correctly classified data instances among the total number of data instances
    \item Precision \citep{NLP4IF-2021-COVID19-task} - Indicates the proportion of correctly classified positive data instances over the total number of data instances classified as positive samples, and computed as follows:
    \begin{equation}
    \small
        Precision = \frac{True\:Positive}{True\:Positive + False\:Positive}
    \end{equation}
    \item Recall \citep{NLP4IF-2021-COVID19-task} - Indicates the proportion of correctly classified positive data instances over the total number of positive data instances, and computed as follows:
    \begin{equation}
    \small
        Recall = \frac{True\:Positive}{True\:Positive + False\:Negative}
    \end{equation}
    \item F1 Score \citep{alam2021fighting,NLP4IF-2021-COVID19-task} - F1 score is the harmonic mean of precision and recall. The score is computed as follows:
    \begin{equation}
    \small
        F1\:Score = 2 * \frac{Precision*Recall}{Precision + Recall}
    \end{equation}
\end{itemize}

Table \ref{table:verifiable_claim_detection_datasets} shows the evaluation metrics used in the existing verifiable claim detection datasets. It can be observed that precision, recall, and F1 score are widely used for the evaluation with the exception of the CheckThat 2022 dataset \citep{nakov2022overview}. This is due to the fairly balanced nature of the CheckThat 2022 dataset, where the authors recommended reporting only accuracy as the evaluation metric. 

\subsection{Discussion}
\begin{figure}[tbp!]
    \centering
    \begin{subfigure}[b]{0.6\textwidth}
    \includegraphics[width=\textwidth]{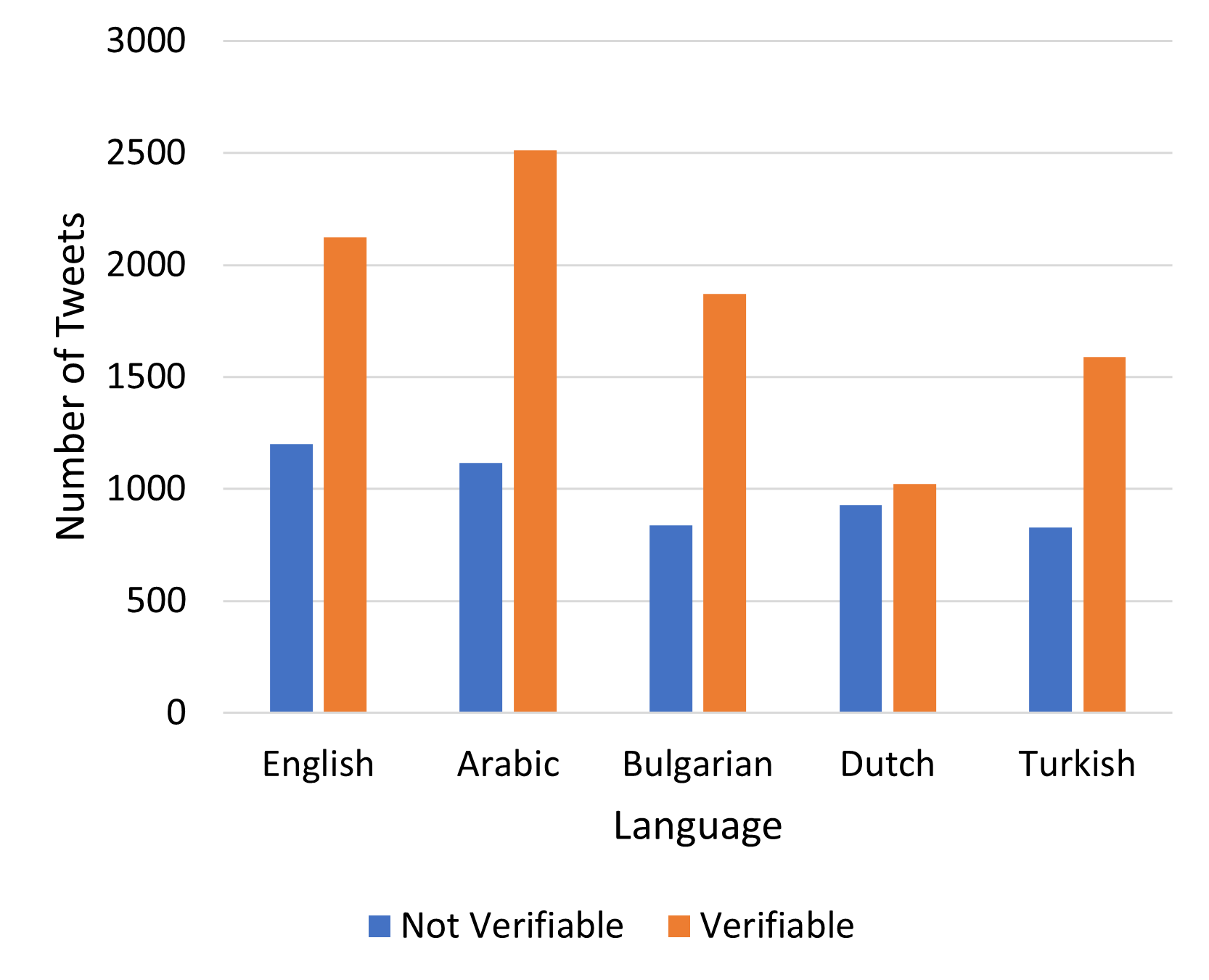}
    \caption{Verifiable Claim Detection}
    \label{fig:CheckThat2022_Verifiable_CD_Dataset_Stat}   
    \end{subfigure}  
    \begin{subfigure}[b]{0.6\textwidth}
    \includegraphics[width=\textwidth]{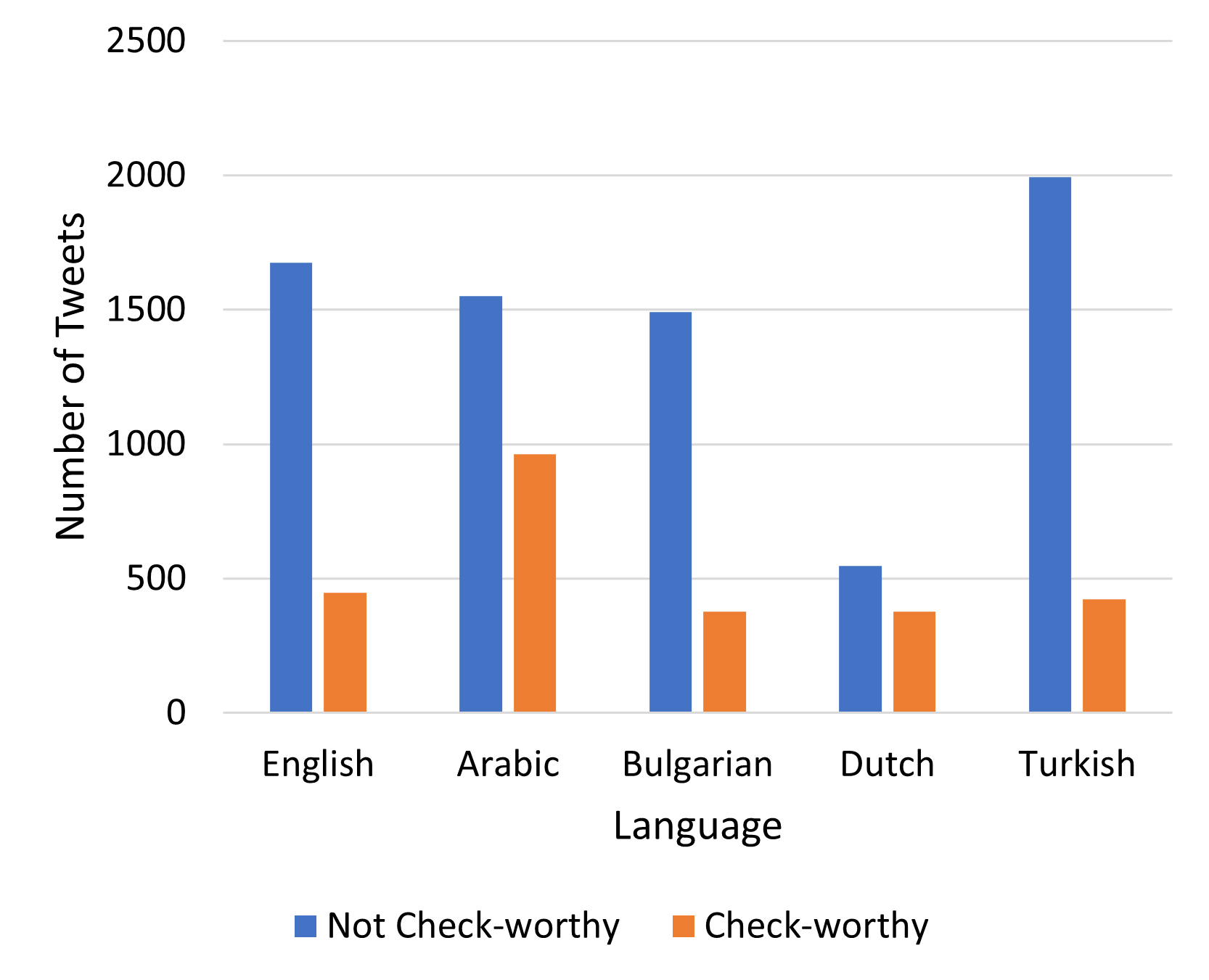}
    \caption{Check-worthy Claim Detection}
    \label{fig:CheckThat2022_CheckWorthy_CD_Dataset_Stat}   
    \end{subfigure}
    \caption{Statistics of CheckThat 2022 verifiable and check-worthy datasets.}
    \label{fig:dataset-stats}
\end{figure}

Table \ref{table:claim_detection_prioritization_summary} summarizes the existing work on verifiable claim detection. Among the three language settings, minimal effort has been made in experimenting with cross-lingual solutions for verifiable claim detection tasks. Moreover, the verifiable claim detection task is often treated with the claim prioritization task using the same solutions. However, these two tasks are different in nature, and require special attention to differentiate between verifiability and priority of a claim. For example, Figure \ref{fig:dataset-stats} shows the statistics of the verifiable and check-worthy claim detection datasets from CheckThat 2020 \citep{nakov2022overview}. Compared to the amount of verifiable claims, relatively very few check-worthy claims are presented in the dataset. This shows the importance of giving attention to the objective of the task to distinguish between verifiability and priority of a claim. 

Further, it can be observed that most of the studies on verifiable claim detection rely on pre-trained transformer models, especially BERT and its variations to obtain state-of-the-art performance with limited fine-tuning. Moreover, limited research has utilized tweet-specific features for the classification of Twitter data, and translation is used as a data augmentation technique to handle the limited training data issue. Further, it can be noticed that performing fine-grained verifiable claim-type identification has been carried out in the literature only in monolingual settings. This could be due to the unavailability of the training data in a multilingual environment and the challenges associated with developing multilingual data annotated with fine-grained claim types. Moreover, very little effort has been made to date in the literature towards utilizing large language models (LLMs) for claim detection. We consider however that employing LLMs for claim detection represents an interesting avenue for future research.           

\section{Claim Prioritization}\label{sec:claim_prioritization}
Prioritizing verifiable claims is a key task in the fact-checking pipeline as not all the verifiable claims can be fact-checked due to a limited availability of resources. However, defining the priority of a verifiable claim is a subjective decision and it depends on multiple factors as discussed in Section \ref{sec:background}. Prioritizing verifiable claims is often treated as a check-worthy claim detection task, which aims at classifying a claim as check-worthy or not. However, the list of claims can be ranked according to the check-worthiness for prioritization. This enables the problem to be solved as either a classification or regression or ranking task. Apart from the check-worthiness, criteria such as attention-worthiness \citep{nakov2022overview}, harmfulness \citep{nakov2022overview,NLP4IF-2021-COVID19-task}, and interest to the general public \citep{NLP4IF-2021-COVID19-task} are used to prioritize the claims. Similar to verifiable claim detection, the prioritization task does not ensure the availability of evidence, and prioritized claims may not have supporting or refusing evidence impeding the determination of a verdict.   

\subsection{Datasets}
A series of CheckThat datasets released since 2018, serve as a rich source of training data for the claim prioritization task in multilingual settings. The initial version of the CheckThat 2018 dataset \citep{nakov2018overview} was released with political debates annotated for ranking check-worthy sentences only in English and Arabic languages. This dataset was collected in the English language, and the Arabic version of it was obtained through manual translation. Later, the problem was tackled as a classification task, and Tweets annotated with various prioritization criteria including check-worthiness, harmfulness, and attention-worthiness were released in multiple languages  \citep{nakov2022overview}. In addition to these criteria, the NLP4IF 2021 dataset includes Tweets annotated with binary labels indicating the \textit{interest to the general public} \citep{NLP4IF-2021-COVID19-task}. One of the largest annotated datasets for check-worthy claim detection was released by the authors of ClaimHunters \citep{beltran2021claimhunter} focusing on languages specific to a region. Table \ref{table:check_worthy_datasets} presents the multilingual claim prioritization datasets.    

\begin{table}
\caption{Multilingual claim prioritization datasets.}
\centering
\label{table:check_worthy_datasets}
\scriptsize
\hspace*{-8em}%Added due to space issue
\begin{tabular}{lllllllll} \toprule
   Dataset  & Criteria & Task  &Label & Language & Topic  & Source & Size & Evaluation \\ \midrule
\begin{tabular}[c]{@{}l@{}}CheckThat 2018\\ \citep{nakov2018overview} \end{tabular}   & Check-worthy & Ranking &  0-1     & \begin{tabular}[c]{@{}l@{}}English\\ Arabic \end{tabular}                                    & Politics & \begin{tabular}[c]{@{}l@{}}Political\\Debate\end{tabular} &  7K - 9K   &  \begin{tabular}[c]{@{}l@{}}MAP\\MRR\\MAP@K\end{tabular}  \\ \midrule
\begin{tabular}[c]{@{}l@{}}NLP4IF 2021\\ \citep{NLP4IF-2021-COVID19-task} \end{tabular}                & \begin{tabular}[c]{@{}l@{}}Interesting\\Harmful\\Requires Attention\end{tabular}         & Classification & \begin{tabular}[c]{@{}l@{}}Yes\\ No\end{tabular}            & \begin{tabular}[c]{@{}l@{}}English\\ Arabic   \\ Bulgarian\end{tabular} & Covid-19                            & Twitter         & 1.3K – 4K   &  \begin{tabular}[c]{@{}l@{}}Precision\\Recall\\F1 Score\end{tabular} \\ \midrule
\begin{tabular}[c]{@{}l@{}}ClaimHunter\\ \citep{beltran2021claimhunter} \end{tabular} & Check-worthy & Classification & \begin{tabular}[c]{@{}l@{}}Yes\\ No\end{tabular} & \begin{tabular}[c]{@{}l@{}}Spanish\\Catalan\\Galician\\Basque\end{tabular}                 & Politics & Twitter & 30K & \begin{tabular}[c]{@{}l@{}}Precision\\Recall\\F1 Score\end{tabular} \\ \midrule
\begin{tabular}[c]{@{}l@{}}CheckThat 2022\\ \citep{nakov2022overview} \end{tabular} & \begin{tabular}[c]{@{}l@{}}Check-worthy\\Harmful\\Attention-worthy\end{tabular}  & Classification & \begin{tabular}[c]{@{}l@{}}Yes\\ No\end{tabular} & \begin{tabular}[c]{@{}l@{}}English\\Arabic\\ Bulgarian\\Spanish\\ Turkish\\Dutch\end{tabular}                 & \begin{tabular}[c]{@{}l@{}}COVID-19\\ Politics\end{tabular} & Twitter & 4K - 6K & F1 Score\\ \bottomrule
\end{tabular}
\end{table}

\subsection{Cross-lingual Claim Prioritization}
ClaimRank \citep{jaradat2018claimrank} is one of the earliest attempts at detecting the check-worthiness of claims in a cross-lingual setting. The authors trained a neural network classifier and input cross-lingual embedding representation of the text along with a wide range of contextual features. This includes TF-IDF scores, part-of-speech (POS) tags, named entities, and topic vectors learned using the topic modeling technique, Latent Dirichlet Allocation (LDA) \citep{lda}. Finally, the likelihood of the classifier was used to rank the claims.  

Similar to verifiable claim detection, multilingual BERT (mBERT) \citep{bert} has been widely used as a solution for claim prioritization in cross-lingual settings. \cite{uyangodage2021can} experimented with both cross-lingual and mono-lingual settings using BERT variations. The authors fine-tuned publicly available language-specific BERT models and mBERT using the language-specific training data and merged training data respectively. They observed similar or improved performance in cross-lingual settings in the CheckThat 2021 dataset. \cite{zengin2021tobb} analyzed a similar approach with language-specific BERT models and mBERT. The authors additionally explored data augmentation techniques such as machine translation and under-sampling (decreasing the size of majority classes) to overcome the imbalanced nature of the dataset. However, none of these techniques increased the performance of check-worthy claim identification. 

\cite{hasanain2022cross} and \cite{kartal2022re} analyzed the zero-shot learning by training the mBERT model only in training data of one language and testing its generalization capability in other languages. The authors observed that mBERT performs as good as the monolingual models trained on target languages. Similar studies have been conducted in the literature using other multilingual models such as XLM-R \citep{beltran2021claimhunter}. \cite{schlicht2023multilingual} proposed to modify the architecture of transformer models by introducing a world language adaptor, a lightweight and modular neural network on top of the multilingual transformers. During the experiments, the authors observed that adaptors trained for world languages were capable of transferring knowledge across languages. 

Multi-tasking was experimented as an alternative solution to improve the performance of claim prioritization in cross-lingual settings. \cite{schlicht2021upv} performed multi-task learning by jointly detecting the language and check-worthy claims. The authors used Sentence BERT \citep{sentence_bert} trained on a multilingual dataset with dedicated fully connected layers for each task. \cite{du2022nus} extended this work by performing a wide range of auxiliary tasks to enhance the performance, and observed an increase in performance for check-worthy claim detection tasks. Notable auxiliary tasks jointly learned include translation to English, verifiable claim detection, harmful tweet detection, and attention-worthy tweet detection. 

\subsection{Multilingual Claim Prioritization}
Similar to the verifiable claim detection in a multilingual setting, a straightforward solution to identify claim priority in the multilingual environment can be achieved by training multilingual pre-trained models such as mBERT \citep{hasanain2020bigir,tarannum2022z,sadouk2023vrai}, XLM-r \citep{tarannum2022z,sadouk2023vrai,aziz2023csecu}, and GPT-3 \citep{sadouk2023vrai} in language-specific training data. Different from these approaches, \cite{nakov2021second} translated the text written in other languages to English first using Google Translation API, and then trained a Support Vector Machine (SVM) classifier \citep{suthaharan2016support} for classifying check-worthy tweets in English. However, the effectiveness of this method was highly dependent on the precision of machine translation. 

\subsection{Monolingual Claim Prioritization}
Developing a monolingual classification of claim priority was often solved by fine-tuning language-specific pre-trained models \citep{williams2021accenture,zhou2021fight} or combining other classification approaches such as neural networks \citep{martinez2020nlp,rony2020claimviz,dutta2022multilingual}, and SVM \citep{cheema2020check} with monolingual word embedding representation. However, the problem can also be approached as a regression task and Logistic Regression \citep{lavalley2008logistic} has been widely used with the word embedding representation for ranking claim priority \citep{kartal2020tobb,kartal2020too}.

The latest attention in this direction has been given to identifying domain-specific claims priority. \cite{pathak2021assessing, pathak2020self} developed solutions specific to news articles based on the assumption that sentences that could well represent the headlines are more check-worthy, and experimented with unsupervised techniques to prioritize check-worthy sentences. \cite{gollapalli2023identifying} attempted to extract medical claims and claim types discussing prevention, diagnoses, cures, treatments, and risks. The authors finetuned the Text-to-Text Transfer Transformer (T5) model \citep{raffel2020exploring} for identifying claim priority, and the BART \citep{bart} model was used to detect the claim types in a zero-shot setting. 

\subsection{Evaluation} \label{sec:claim_prioritization_evaluation}
Evaluation of claim prioritization depends on the nature of the task. The following measures are used in the literature to evaluate the ranking tasks.
\begin{itemize}
    \item MAP@K (Mean Average Precision @ K) \citep{nakov2022overview} - Computed as the mean of the average precision value of all the data instances. Here, the average precision is computed as the average precision score for the range of K value (average of precision @ 1 to precision @ K). 
    \item MRR (Mean Reciprocal Ranking) \citep{shaar2020known,kazemi2021claim} - Given the actual ranking of top K elements, MRR is calculated using their corresponding retrieved as follows,
    \begin{equation}
        MRR = \frac{1}{K}\sum_{i=1}^{K}\frac{1}{rank_K}
    \end{equation}
\end{itemize}

Metrics such as accuracy, precision, recall, and F1-score (refer Section \ref{sec:verifiable-claim-detection-evaluation}) are widely used as an evaluation measure when the prioritization is carried out as a classification task. Table \ref{table:check_worthy_datasets} shows the evaluation metrics used in the existing claim prioritization datasets. It can be observed that a wide selection of metrics is used in the literature depending on whether the task is modeled as a classification or ranking problem. It is worth noting that the CheckThat 2022 dataset \citep{nakov2022overview} recommended reporting the F1 score to account for class imbalance.

\subsection{Discussion}
Prioritizing a claim is generally modeled as an estimation of the check-worthiness of a claim, and the problem is solved as either a classification or ranking task in the literature. Table \ref{table:claim_detection_prioritization_summary} summarizes the existing work on claim prioritization. It can be observed that transformer-based models are widely used in all three settings, and various data augmentation techniques including up-sampling, down-sampling, and machine translation were used to overcome the imbalance in the training dataset. Further, multi-tasking is also experimented as a solution to transfer the language and task knowledge in cross-lingual settings. Similar to the trend of treating both verifiable and check-worthy claims with the same solutions, different prioritization tasks are treated the same without giving much attention to the actual objective (e.g. check-worthiness, attention-worthiness, and harmfulness). Possibly this could be an interesting future direction for developing prioritization solutions incorporating the actual priority criteria.  

\begin{table}[tbp!]
\vspace*{-6em}%Added due to space issue
\caption{Summary of existing claim detection and prioritization research.}
\centering
\label{table:claim_detection_prioritization_summary}
\scriptsize
\hspace*{-7em}%Added due to space issue
\begin{tabular}{p{6cm} l l l l l l l l l l l l l} \toprule
                                                  Research & \multicolumn{2}{c}{Objective}                              & \multicolumn{3}{c}{Language}                                                      & \multicolumn{1}{c}{Model} & \multicolumn{2}{c}{\begin{tabular}[c]{@{}c@{}}Data \\ Augmentation\end{tabular}}                 & \multicolumn{1}{c}{Input} & \multicolumn{1}{c}{Learning} & \multicolumn{3}{c}{Dataset}                                                       \\ \midrule
                                                  & \rotatebox{90}{Verifiability}                & \rotatebox{90}{Priority}              & \rotatebox{90}{Cross-lingual}             & \rotatebox{90}{Multi-lingual}             & \rotatebox{90}{Mono-lingual}              & \rotatebox{90}{Transformer}               & \rotatebox{90}{Machine Translation}       & \rotatebox{90}{Sampling}                  & \rotatebox{90}{Twitter features}          & \rotatebox{90}{Multi-tasking}                & \rotatebox{90}{CheckThat}                 & \rotatebox{90}{NLP4IF}                   & \rotatebox{90}{Other}                     \\ \midrule
\cite{panda2021detecting}         & \checkmark  & \checkmark  & \checkmark  & \checkmark  & - & \checkmark  & - & - & - & -    & - & \checkmark  & - \\
\cite{uyangodage2021can}          & \checkmark  & \checkmark  & \checkmark  &                           & \checkmark  & \checkmark  & - & - & - & -    & \checkmark  & \checkmark  & - \\
\cite{alam2020fighting}           & \checkmark  & \checkmark  & \checkmark  &                           & \checkmark  & \checkmark  & - & - & \checkmark  & -    & \_                        & \checkmark  & \_                        \\
\cite{agrestia2022polimi}         & \checkmark  & \checkmark  & \checkmark  & - & - & \checkmark  & - & - & - & -    & \checkmark  & - & - \\
\cite{husunbeyi2022rub}           & \checkmark  & \checkmark  & - & \checkmark  & \checkmark  & \checkmark  & - & - & - & -    & \checkmark  & - & - \\
\cite{savchev2022ai,eyuboglu2023fight}              & \checkmark  & \checkmark  & - & \checkmark  & \checkmark  & \checkmark  & \checkmark  & - & - & -    & \checkmark  & - & - \\
\cite{suri2022asatya}             & \checkmark  & \checkmark  & - & - & \checkmark  & \checkmark  & \checkmark  & - & \checkmark  & -    & \checkmark  & - & - \\
\cite{henia2021icompass,hussein2021damascusteam}          & \checkmark  & \checkmark  & - & - & \checkmark  & \checkmark  & - & - & - & -    & - & \checkmark  & - \\
\cite{alam2021fighting,smeros2021sciclops}           & \checkmark  & \checkmark  & - & - & \checkmark  & \checkmark  & - & - & - & -    & - & - & \checkmark  \\
\cite{woloszyn2021towards,prabhakar2020claim}        & \checkmark  & - & - & - & \checkmark  & \checkmark  & - & - & - & -    & - & - & \checkmark  \\
\cite{konstantinovskiy2021toward} & \checkmark  & - & - & - & \checkmark  & - & - & - & - & -    & - & - & \checkmark  \\
\cite{zengin2021tobb}             & - & \checkmark  & \checkmark  &                           & \checkmark  & \checkmark  & \checkmark  & \checkmark  & - & -    & \checkmark  & - & - \\
\cite{hasanain2022cross,kartal2022re}          & - & \checkmark  & \checkmark  &                           & \checkmark  & \checkmark  & - & - & - & -    & \checkmark  & - & - \\
\cite{schlicht2021upv}            & - & \checkmark  & \checkmark  & - & - & \checkmark  & - & - & - & \checkmark     & \checkmark  & - & - \\
\cite{du2022nus}                  & - & \checkmark  & \checkmark  & - & - & \checkmark  & - & \checkmark  & \checkmark  & \checkmark     & \checkmark  & - & - \\
\cite{beltran2021claimhunter}     & - & \checkmark  & \checkmark  & - & - & \checkmark  & - & - & - & -    & - & - & \checkmark  \\
\cite{schlicht2023multilingual}   & - & \checkmark  & \checkmark  & - & - & \checkmark  & - & - & - & -    & \checkmark  & - & - \\
\cite{jaradat2018claimrank}       & - & \checkmark  & \checkmark  & - & - & - & - & - & - & -    & \checkmark  & - & - \\
\cite{sadouk2023vrai}             & - & \checkmark  & - & \checkmark  & \checkmark  & \checkmark  & - & \checkmark  & - & -    & \checkmark  & - & - \\
\cite{aziz2023csecu}              & - & \checkmark  & - & \checkmark  & \checkmark  & \checkmark  & - & - & - & -    & \checkmark  & - & - \\
\cite{hasanain2020bigir}          & - & \checkmark  & - & \checkmark  & - & \checkmark  & - & - & \checkmark  & -    & \checkmark  & - & - \\
\cite{tarannum2022z}              & - & \checkmark  & - & \checkmark  & - & \checkmark  & - & \checkmark  & - & -    & \checkmark  & - & - \\
\cite{nakov2021second}            & - & \checkmark  & - & \checkmark  & - & - & \checkmark  & - & - & -    & - & - & \checkmark  \\
\cite{cheema2020check,williams2021accenture,kartal2020tobb,kartal2020too}            & - & \checkmark  & - & - & \checkmark  & \checkmark  & - & - & - & -    & \checkmark  & - & - \\
\cite{pathak2021assessing,gollapalli2023identifying}        & - & \checkmark  & - & - & \checkmark  & \checkmark  & - & - & - & -    & - & - & \checkmark  \\
\cite{martinez2020nlp,zhou2021fight}            & - & \checkmark  & - & - & \checkmark  & - & - & \checkmark  & - & -    & \checkmark  & - & - \\
\cite{rony2020claimviz}           & - & \checkmark  & - & - & \checkmark  & - & - & - & - & -    & - & - & \checkmark \\ \bottomrule
\end{tabular}
\end{table}

% \usepackage{multirow}
\begin{table}[tbp!]
\caption{Summary of models used for claim detection and prioritization research.}
\centering
\label{table:claim_detection_prioritization_model_summary}
\scriptsize
\hspace*{-7em}%Added due to space issue
\begin{tabular}{llp{8cm}}\toprule
Model Type                           & Model                  & Reference                                                                                                                                                                                                                                                                                     \\ \midrule
\multirow{3}{*}{Machine Learning}    & Logistic Regression    & \cite{kartal2020tobb,panda2021detecting,konstantinovskiy2021toward,beltran2021claimhunter}                                                                                                                                                                                                           \\
                                     & Random Forest          & \cite{smeros2021sciclops,tarannum2022z}                                                                                                                                                                                                                                                             \\
                                     & Support Vector Machine & \cite{konstantinovskiy2021toward,schlicht2021upv,beltran2021claimhunter,tarannum2022z}                                                                                                                                                                                                            \\ \midrule
\multirow{5}{*}{Deep Learning}       & LSTM                   & \cite{prabhakar2020claim}                                                                                                                                                                                                                                                                            \\
                                     & Bi-LSTM                & \cite{martinez2020nlp,rony2020claimviz}                                                                                                                                                                                                                                                              \\
                                     & CNN                    & \cite{martinez2020nlp}                                                                                                                                                                                                                                                                               \\
                                     & Feed Forward NN        & \cite{martinez2020nlp,konstantinovskiy2021toward}                                                                                                                                                                                                                                                    \\
                                     & Falir                  & \cite{woloszyn2021towards}                                                                                                                                                                                                                                                                           \\ \midrule
\multirow{17}{*}{Transformer Family} & BERT                   & \cite{kartal2020tobb,prabhakar2020claim,cheema2020check,alam2020fighting,panda2021detecting,pathak2021assessing,zhou2021fight,  uyangodage2021can,smeros2021sciclops,alam2021fighting,zengin2021tobb,hasanain2022cross,savchev2022ai,kartal2022re,suri2022asatya,eyuboglu2023fight,sadouk2023vrai} \\
                                     & mBERT                  & \cite{hasanain2020bigir,alam2020fighting,panda2021detecting,uyangodage2021can,alam2021fighting,zengin2021tobb,hasanain2022cross,kartal2022re,tarannum2022z,schlicht2023multilingual,sadouk2023vrai}                                                                                               \\
                                     & XLM-r                  & \cite{alam2020fighting,alam2021fighting,beltran2021claimhunter,husunbeyi2022rub,tarannum2022z,du2022nus,schlicht2023multilingual,aziz2023csecu}                                                                                                                                                   \\
                                     & RoBERTa                & \cite{alam2020fighting,zhou2021fight,alam2021fighting,woloszyn2021towards,williams2021accenture,savchev2022ai,sadouk2023vrai,eyuboglu2023fight}                                                                                                                                                   \\
                                     & DistilBERT             & \cite{prabhakar2020claim,savchev2022ai,eyuboglu2023fight}                                                                                                                                                                                                                                            \\
                                     & BERTweet               & \cite{zhou2021fight,woloszyn2021towards,husunbeyi2022rub,aziz2023csecu}                                                                                                                                                                                                                              \\
                                     & SentenceBERT           & \cite{schlicht2021upv}                                                                                                                                                                                                                                                                               \\
                                     & ConvBERT               & \cite{husunbeyi2022rub}                                                                                                                                                                                                                                                                              \\
                                     & ALBERT                 & \cite{alam2021fighting,eyuboglu2023fight,sadouk2023vrai}                                                                                                                                                                                                                                             \\
                                     & XL-Net                 & \cite{kartal2022re,sadouk2023vrai}                                                                                                                                                                                                                                                                   \\
                                     & Transformer            & \cite{panda2021detecting,du2022nus,gollapalli2023identifying}                                                                                                                                                                                                                                        \\
                                     & FastText               & \cite{kartal2020tobb,alam2021fighting}                                                                                                                                                                                                                                                               \\
                                     & AraBERT                & \cite{alam2020fighting,hussein2021damascusteam,henia2021icompass,zengin2021tobb,hasanain2022cross,aziz2023csecu,eyuboglu2023fight}                                                                                                                                                                \\
                                     & Spanish BERT           & \cite{uyangodage2021can}                                                                                                                                                                                                                                                                             \\
                                     & BERTurk                & \cite{uyangodage2021can,zengin2021tobb,hasanain2022cross,husunbeyi2022rub,eyuboglu2023fight}                                                                                                                                                                                                         \\
                                     & DutchBERT              & \cite{alam2020fighting}                                                                                                                                                                                                                                                                              \\
                                     & SalvicBERT             & \cite{hasanain2022cross}                                                                                                                                                                                                                                                                             \\ \midrule
Large Language Models                & GPT                    & \cite{agrestia2022polimi,sadouk2023vrai}  \\ \bottomrule                                                                                                                                                                                                                                           
\end{tabular}
\end{table}

Table \ref{table:claim_detection_prioritization_model_summary} summarizes the models used in the literature for claim detection and claim prioritization. It can be noticed that transformer-based models are widely used compared to machine learning approaches and other deep learning learning approaches. Especially the BERT model and its architectural variations such as RoBERTa, XLM-r, ALBERT, BERTweet, Sentence BERT, DistilBERT, and ConvBERT, its multilingual variations such as mBERT, and the language-specific variations such as AraBERT, Spanish BERT, and BERTurk are commonly applied. Further, it can be noticed that large language models are yet to be explored in this line of research.

\section{Claim Matching}\label{sec:claim matching}
Claim matching is the task of identifying a pair of claims that can be addressed with the same fact-check \citep{kazemi2021claim}. This can be handled as either a classification task to classify whether the two claims match or not, or a regression or semantic similarity task to generate a score indicating the strength of the match. When modeling as a classification problem, the likelihood of the classifier can also be used as a score indicating the probability of the two statements discussing the same claim. The task can be further extended as a search problem from a database of verified claims, by producing a ranked list of verified claims matching the input claim using the scores obtained via classification, or regression, or semantic similarity function. This extended task is referred to as \textit{fact-checked claim retrieval} or \textit{verified claim retrieval}. Interestingly, the claim matching task is highly related to the next component of the fact-checking pipeline; evidence retrieval. Here, the underlying idea of finding the relationship between two claims or a claim-evidence pair remains the same, and two or more claims matched to the same evidence can be treated as similar claims that can be fact-checked together. 

\subsection{Datasets}
As mentioned previously, claim-matching tasks are treated with various objectives in the literature, and a wide range of datasets serving these objectives are available in multilingual environments. This includes matching two tweets \citep{kazemi-etal-2021-claim}, matching tweets with a report \citep{kazemi2022matching}, and also matching a verified claim with tweets or social media posts \citep{shaar2021overview,nielsen2022mumin,pikuliak2023multilingual}. Table \ref{table:claim_matching_datasets} summarizes the existing multilingual claim-matching datasets. 

\begin{table}[tbp!]
\caption{Multilingual claim matching datasets.}
\centering
\label{table:claim_matching_datasets}
\tiny
\hspace*{-12em}%Added due to space issue
\begin{tabular}{llllllll}
\toprule
\textbf{Dataset}     & \textbf{Label}                                                        & \textbf{Language}                                                                              & \textbf{Language Pair}                                              & \textbf{Topic}                                                       & \textbf{Source}                                                 & \textbf{Size}     & Evaluation\\ \midrule
\cite{kazemi-etal-2021-claim}  & Claim Pairs                                                  & \begin{tabular}[c]{@{}l@{}}English\\ Hindi\\ Bengali\\ Malayalam\\ Tamil\end{tabular} & Monolingual                                                         & \begin{tabular}[c]{@{}l@{}}Covid-19\\ Politics\end{tabular} & WhatsApp                                                        & 300 - 650 Pairs       & \begin{tabular}[c]{@{}l@{}}Accuracy\\Precision\\Recall\\F1 Score\end{tabular} \\ \midrule
\cite{shaar2021overview} & \begin{tabular}[c]{@{}l@{}}Claim-Tweets\\ Pairs\end{tabular} & \begin{tabular}[c]{@{}l@{}}English\\Arabic\end{tabular}        & Monolingual & Multitopic                                                          & \begin{tabular}[c]{@{}l@{}}Twitter\\Snopes\\AraFact \citep{ali2021arafacts}\\ClaimsKG\\ \citep{tchechmedjiev2019claimskg}\end{tabular} & 2.5K Pairs & \begin{tabular}[c]{@{}l@{}}MRR\\MAP@K\\Precision@K\end{tabular}\\ \midrule
\cite{kazemi2022matching} & \begin{tabular}[c]{@{}l@{}}Claim-Report\\ Pairs\end{tabular} & \begin{tabular}[c]{@{}l@{}}English\\ Hindi\\ Spanish\\ Portuguese\end{tabular}        & \begin{tabular}[c]{@{}l@{}}Monolingual\\ Hindi-English\end{tabular} & Multitopic                                                          & \begin{tabular}[c]{@{}l@{}}Twitter\\Google Fact Check Tools\end{tabular} & 400 - 4.8K Pairs &  \begin{tabular}[c]{@{}l@{}}Accuracy\\F1 Score\\MRR\\MAP@K\end{tabular}\\ \midrule
\begin{tabular}[c]{@{}l@{}}MuMiN\\ \citep{nielsen2022mumin}\end{tabular} & \begin{tabular}[c]{@{}l@{}}Claim-Tweet\\ Pairs\end{tabular} & 41 Languages & Multilingual & Multitopic                                                          & \begin{tabular}[c]{@{}l@{}}Twitter\\Google Fact Check Tools\end{tabular} & \begin{tabular}[c]{@{}l@{}}13K Claims\\ 21M Tweets\end{tabular} &  F1 Score \\\midrule
\begin{tabular}[c]{@{}l@{}}MultiClaim\\ \citep{pikuliak2023multilingual}\end{tabular} & \begin{tabular}[c]{@{}l@{}}Claim-Post\\ Pairs\end{tabular} & 27 Languages & Multilingual & Multitopic                                                          & \begin{tabular}[c]{@{}l@{}}Face book, Twitter\\Instragam\\Google Fact Check Tools\end{tabular} & 31K Pairs 
 &  \begin{tabular}[c]{@{}l@{}}Precision@K\\Recall@K\end{tabular}\\\midrule
\begin{tabular}[c]{@{}l@{}}MMTweets\\ \citep{singh2023finding}\end{tabular} & \begin{tabular}[c]{@{}l@{}}Claim-Misinformation\\Tweet\\ Pairs\end{tabular} & \begin{tabular}[c]{@{}l@{}}English\\ Hindi\\ Spanish\\ Portuguese\end{tabular}  & Multilingual & Covid-19                                                          & \begin{tabular}[c]{@{}l@{}}Twitter\\Fact-checking Organizations\end{tabular} & 1.6K Pairs &  \begin{tabular}[c]{@{}l@{}}MRR\\MAP@K\end{tabular} \\\bottomrule
\end{tabular}
\end{table}

\subsection{Cross-lingual Claim Matching}
Very little effort has been made in the direction of experimenting with claim matching in cross-lingual settings. The authors of the \textit{MultiClaim} dataset \citep{pikuliak2023multilingual} used various multilingual and monolingual embedding representations of posts and claims, and the distance between the vector representation was used as a similarity function to retrieve fact-checked claims. Further, they compared the embedding-based retrieval using the BM25 ranking algorithm \citep{BM25}. The experiment results showed the BM25 algorithm is ineffective in handling multilingual environments and multilingual embedding representations such as LaBSE \citep{feng2022language}  retrieve similar claims effectively. Instead of utilizing the embeddings directly, the authors of \textit{MMTweets} dataset \citep{singh2023finding} fine-tuned various multilingual models mBERT \citep{bert}, XLM-r \citep{xlm-r} and LaBSE \citep{feng2022language} to tackle the problem as a classification task. They observed a similar trend of low performance of the BM25 algorithm in multilingual settings, and LaBSE has consistently been the best model. Following these observations, fine-tuning LaBSE \citep{feng2022language} was used as an effective solution for claim matching in the literature \citep{nielsen2022mumin}.    

Different from these approaches, \cite{larraz2023semantic} integrated both a semantic similarity-based technique and classification to perform cross-lingual claim matching for political discourse. The K-Nearest Neighbors (KNN) algorithm \citep{peterson2009k} was used to find the top N neighbors of the input claim first, and finally, a BERT-based classifier was employed to classify the claim pairs. The authors observed that the XLM-r \citep{xlm-r} model was performing better among the BERT-based models compared to identifying the claim pairs among the top N neighbors. 

\cite{kazemi2022matching} solved a similar problem of matching a tweet to a report containing a similar claim as both classification and ranking tasks. The authors fine-tuned the multilingual transformer model XLM-r \citep{xlm-r} to solve the classification problem. Ranking similar reports for the given tweet was achieved by developing a word embedding-based similarity search system using the sentence embedding representations \citep{sentence_bert,feng2022language}. 

\subsection{Multi-lingual Claim Matching}
Interestingly, the claim matching problem is widely experimented in either cross-lingual settings or monolingual settings, and limited effort has been made to develop multi-lingual claim detection. In the only such effort to date, \cite{kazemi2021claim} solved claim matching as a classification problem in a multilingual environment, by training XLM-r \citep{xlm-r} model for each monolingual language pair in the dataset they released, and then applying the BM25 ranking algorithm on language-specific embedding representation of claims. The authors observed an increase in performance in claim matching, when combining BM25 with XLM-r embedding representation, compared to performing BM25 ranking independently. 

\subsection{Mono-lingual Claim Matching}
One of the pioneering works in this direction was experimented by \cite{shaar2020known}. Given an unverified claim, the authors retrieved a ranked list of fact-checked claims. Both unverified claims and verified claims were represented using the sentence embedding representation obtained via the BERT model \citep{bert} and its variations. Cosine similarity between the sentence embedding representations was used to retrieve a set of matching verified claims. Finally, a Support Vector Machine (SVM) model \citep{suthaharan2016support} was trained to rank the retrieved claim list. \cite{shaar2022assisting} applied the same technique with a wide range of sentence embedding representations combined with the SVM model for ranking. Apart from the integration of regression models with semantic similarity approaches, recent studies \citep{bouziane2020team,mansour2022did,mansour2023not} have focused on solving the verified claim retrieval problem as a classification task by fine-tuning BERT models.

A more recent study on monolingual claim matching \citep{singh2023utdrm} shows the importance of having a larger scale training data for accurately retrieving the fact-checked claims. The authors generated a large number of synthetic claims from fact-checked claims using text-to-text transfer transformer (T5) and Chat-GPT models.\footnote{https://platform.openai.com/docs/models.} A large-scale synthetic dataset was used to fine-tune transformer models for matching claim pairs. Experiment results show that the proposed unsupervised method yields similar or slightly improved retrieval performance compared to the state-of-the-art transformer models directly trained on the claim-matching training data. This encourages generating synthetic data using large language models at a lower cost of computational efficiency to overcome the challenges associated with annotating large-scale training data.    

\subsection{Evaluation}\label{sec:claim_matching_evaluation}
Evaluating claim matching depends on whether the problem is treated as a classification or ranking task. Evaluation metrics discussed in Section \ref{sec:verifiable-claim-detection-evaluation} for the classification tasks and in Section \ref{sec:claim_prioritization_evaluation} for the ranking tasks can be applied according to the nature of the problem. Table \ref{table:claim_matching_datasets} shows the evaluation metrics used in the existing claim-matching datasets. It can be observed that Mean Reciprocal Ranking (MRR) and Mean Average Precision@K (MAP@K) are widely used for evaluation. Interestingly MultiClaim \citep{pikuliak2023multilingual} reports Precision@K and Recall@K as the ranking measures.  

\subsection{Discussion}
Table \ref{table:claim_clustering_research} summarizes the claim-matching tasks in the literature. This problem is often solved as either a classification or regression or a semantic similarity task in the literature. In all these cases, transformer models are widely used as either the classification model or the embedding representation of the claims due to their superior performance in understanding the language and the task. Few studies experimented with ranking algorithms such as BM25 \citep{BM25} for retrieving similar claims in cross-lingual settings, and the experiment results reveal the algorithms perform at their optimal when combined with embedding representations such as LaBSE \citep{feng2022language}. Further, the difficulties associated with annotating large-scale training data have been addressed through the utilization of large language models for synthetic data generation. 

\section{Claim Clustering}\label{sec:claim clustering}
This section summarizes existing work on the claim clustering task. The objective of clustering claims is to identify a set (including a pair or more) of claims expressing similar claims or similar topics. The former can be seen as a generalized problem of claim matching to identify the set of claims that can be fact-checked together. However, more coarse-grained clustering of claims can also be performed to further analyze claims belonging to the same topic. According to our knowledge, there is no training data annotated for identifying claim clusters existing even for the monolingual setting.

\subsection{Cross-lingual Claim Clustering}
Very little effort has been made in the direction of identifying multi-lingual clusters in the literature. The challenge of this task escalates with the unavailability of datasets annotated for claim clusters and makes it further difficult to evaluate the solutions proposed. \cite{kazemi2021claim} trained an XLM-R transformer model \citep{xlm-r} to obtain the sentence embedding of the claims. Once the embedding representations were obtained, a single-link hierarchical clustering technique was applied to verify the existence of multi-lingual clusters. While the authors observed meaningful multilingual clusters in the dataset, they were enabled to evaluate the accuracy of the clusters obtained. A similar approach was used by \cite{nielsen2022mumin} to cluster the claims using sentence embedding obtained via the pre-trained Sentence BERT model \citep{sentence_bert}. The authors applied the HDBSCAN clustering approach \citep{mcinnes2017hdbscan} to the UMAP projection \citep{mcinnes2018umap} of the sentence embedding. While the authors observed the existence of topic clusters among the claims, they were enabled to evaluate the generated clusters due to the unavailability of the annotated dataset.

\subsection{Monolingual Claim Clustering}
The issue of the non-existence of a relevant dataset persists for monolingual claim clustering as well. \cite{adler2019real} overcame this issue with the assumption that claims pertaining to the same news article should be clustered together. The authors used the Google USE Large pre-trained model \citep{cer2018universal} to obtain the embedding representation of the sentences and applied the DBSCAN clustering technique \citep{ester1996density} to identify the claim clusters. They configured the clustering algorithm to allow even one claim to be a cluster to support sequential clustering (building the clusters dynamically by adding one element at a time). However, this could result in a huge amount of clusters, and the authors resolved this issue by applying the Louvain Community Detection algorithm \citep{blondel2008fast} to the identified clusters to determine the final clusters. The average percentage of claims belonging to the same story within a cluster was reported as a quantitative measure of the accuracy of the clusters identified.  

\cite{smeros2021sciclops} identified scientific claim clusters and reported a modified version of the Silhouette score \citep{ROUSSEEUW198753}, a widely used clustering evaluation metric for unlabelled data as the evaluation metric. The authors applied a wide range of domain-specific techniques to cluster scientific claims and related research papers represented as a bipartite graph. A notable clustering technique includes topic extraction from claim statements using Latent Dirichlet Allocation \citep{lda}, and applying clustering algorithms such as K-Means \citep{lloyd1982least} on topic vectors of claims. A similar idea of applying K-Means clustering techniques on the embedding representation of social media posts was also experimented to cluster posts discussing similar claims \citep{hale2024analyzing}. Even though the authors couldn't report any quantitative measure of the accuracy of the clusters identified, the manual analysis revealed the existence of similar claims across multiple social platforms in different formats, languages, and lengths. This shows the requirement of performing claim clustering for effective fact-checking and misinformation management. 

\subsection{Evaluation}
Due to the unavailability of a claim clustering dataset with cluster annotation, cluster evaluation metrics proposed for unlabeled clusters can be adopted for this task as a quantitative measure. Following is the cluster evaluation metric adopted for the claim clustering task in the literature:
\begin{itemize}
    \item Silhouette score \citep{ROUSSEEUW198753} - Computed as an average score of Silhouette score of all the data points. For a given data point $i$, the score is computed as follows,
    \begin{equation}
        S_i = \frac{b_i - a_i}{max(b_i, a_i)}
    \end{equation}
    where $a_i$ is the average distance of data point $i$ to any other point in the same cluster, and $b_i$ is the average distance of point $i$ to all the other points in the nearest cluster. Here, the distance between two claims can be defined using a semantic similarity function.   
\end{itemize}

\subsection{Discussion}
Table \ref{table:claim_clustering_research} summarizes the claim clustering tasks in the literature. Clustering multilingual claims has been attempted on very limited occasions in the literature, and the quality of the clusters identified could not be precisely measured due to the unavailability of annotated datasets. As an alternative, traditional cluster evaluation metrics used for unlabelled clusters have been reported as a quantitative measure of claim cluster detection. Compared to the claim clustering task, more effort has been devoted to identifying a pair of claims or retrieving a list of claims discussing the same or similar claims. While this task can be treated with different objectives such as grouping similar claims for the same fact-check or grouping claims discussing the same topic, this research direction is still in its infancy and demands a thorough definition of the task and annotated datasets. 

\begin{table}[tbp!]
\caption{Summary of existing claim matching and claim clustering research.}
\centering
\hspace*{-5em}%Added due to space issue
\label{table:claim_clustering_research}
\small
\begin{tabular}{p{8cm}lllllllll}\toprule
                                    Research      & \multicolumn{2}{l}{Task}                              & \multicolumn{2}{l}{Language}                          & Model                     & \multicolumn{4}{l}{Technique}                                                     \\ \midrule
                                          & \rotatebox{90}{Claim Matching}                & \rotatebox{90}{Clustering}                  & \rotatebox{90}{Cross-lingual}             & \rotatebox{90}{Monolingual}               & \rotatebox{90}{Transformer}               & \rotatebox{90}{Similarity Function}       & \rotatebox{90}{Classification}            & \rotatebox{90}{Regression}                &
                                          \rotatebox{90}{Clustering Algorithm}                \\ \midrule
\cite{kazemi2021claim}                           & \checkmark  & \checkmark  & \checkmark  & - & \checkmark  & \checkmark  & \checkmark  & - & \checkmark \\
\cite{nielsen2022mumin}                          & \checkmark  & \checkmark & \checkmark  & - & \checkmark  & - & \checkmark & - & \checkmark \\
\cite{kazemi2022matching,larraz2023semantic}						  & \checkmark & -  & \checkmark  & - & \checkmark  & \checkmark  & \checkmark  & - & - \\
\cite{pikuliak2023multilingual}                  & \checkmark & -  & \checkmark  & \checkmark  & \checkmark  & \checkmark  & - & - & - \\
\cite{singh2023finding}                          & \checkmark & -  & \checkmark  & - & \checkmark  & - & \checkmark  & - & - \\
\cite{shaar2022assisting}                        & \checkmark & -  & - & \checkmark  & \checkmark  & \checkmark  & - & - & - \\
\cite{shaar2020known}                            & \checkmark & -  & - & \checkmark  & \checkmark  & \checkmark  & - & \checkmark  & - \\
\cite{bouziane2020team,mansour2022did,singh2023utdrm,mansour2023not}   						  & \checkmark & -  & - & \checkmark  & \checkmark  & - & \checkmark  & - & - \\ 
\cite{adler2019real,hale2024analyzing}                             & -  & \checkmark & - & \checkmark  & \checkmark  & - & - & - & \checkmark \\
\cite{smeros2021sciclops}                        & -  & \checkmark & - & \checkmark  & - & - & - & - & \checkmark \\ \bottomrule
\end{tabular}
\end{table}

\section{Open Challenges}\label{sec:challenges}
This section discusses open challenges associated with the ongoing research on multilingual claim detection.
\begin{itemize}
    \item \textbf{Limited Multilingual Datasets:} One of the key aspects hindering the progress of multilingual claim detection research is the unavailability of training data. Especially, comprehensive multitopic claim detection datasets, verifiable claim type detection datasets, claim clustering datasets, and explainable claim detection are yet to be developed for the research progress even in monolingual settings. Automated generative approaches may be used as an alternative to generating claim detection datasets \citep{bussotti2023generation,veltri2023data}. 
    \item \textbf{Validity of Data Sources:} Annotating a massive amount of factual statements for their verifiability, priority, and similarity is a tedious and expensive task. This resulted in relying on existing tools such as the Google Fact Checking tool to partially automate the creation of training datasets. Further, the transparency in the data gathering and annotation process often does not persist, and these factors question the credibility of the dataset as well as the solutions developed on it.
    \item \textbf{Consolidated Definition of the Tasks:} Defining verifiability, priority, and similarity of claims may depend on various factors such as source, topic, target audience, etc. Therefore, a wide range of definitions are used in the literature to tackle all three aspects of the claim detection problem. This highly hinders the research progress with unified agreement on the definition of the tasks. 
    \item \textbf{Change of Claim Status with Time:} Both the true value and the requirement to determine the verifiability, priority, and similarity of claims may change over time. Further, incorporating this temporal nature of the problem is scarcely explored in the literature, mainly due to the unavailability of datasets meeting these objectives, and the challenges associated with simulating the real-time environment for accurate experiments.
    \item \textbf{Demand of Generalizable Solutions:} As previously discussed, the source of factual statements can be from various platforms and can be articulated in various formats, languages, and modalities. Recent studies \citep{hale2024analyzing} have shown evidence of the existence of the same claims across multiple platforms, written in multiple formats, lengths, and details. While this demands more generalizable solutions to identify claims regardless of these factors, most of the existing research focuses on developing solutions specific to a source, data format, language, and modality.
    \item \textbf{Language Imbalance in Datasets:} Most of the existing multilingual datasets are composed of a higher number of annotated samples for high-resource languages such as English, compared to lesser-resourced languages. While existing research tried to tackle this problem via sampling, and via augmenting data in the underrepresented languages through machine translation techniques, this could lead to biases in the training when the model is provided with more data on certain languages.
\end{itemize}

\section{Conclusion}\label{sec:conclusion}
Within the timely research area of automated fact-checking, this survey presents a comprehensive review of the existing multilingual claim detection research. Specifically, we detail the state-of-the-art techniques used to identify the verifiability, priority, and similarity of multilingual claims in the literature. It can be observed that relatively more effort has been given to prioritizing claims compared to identifying verifiable claims and similar claims. Further, fine-tuning multilingual pre-trained transformer models is widely used as a solution in all three problems, due to their powerful nature in transferring knowledge across languages and tasks. 

As discussed previously, various factors affect the progress of multilingual claim detection research. Notably, the unavailability of datasets meeting the task requirement serves as a key challenge. Promising future direction includes the development of credible and comprehensive datasets, generalized solutions, explainable claim detection, time-aware claim detection, and prompt-based fine-tuning using large language models.    

\section*{Declaration of competing interest}
The authors declare that they have no known competing financial interests or personal relationships that could have appeared to influence the work reported in this paper.

\section*{Acknowledgments}
This project is funded by the European Union and UK Research and Innovation under Grant No. 101073351 as part of Marie Skłodowska-Curie Actions (MSCA Hybrid Intelligence to monitor, promote, and analyze transformations in good democracy practices).



\bibliographystyle{elsarticle-harv} 
\bibliography{references}

\end{document}